\documentclass[runningheads]{llncs}

\usepackage{eccv}

\RequirePackage{xspace}
\makeatletter
\DeclareRobustCommand\onedot{\futurelet\@let@token\@onedot}
\def\@onedot{\ifx\@let@token.\else.\null\fi\xspace}

\def\eg{\emph{e.g}\onedot} 

\def\ie{\emph{i.e}\onedot}

\def\etal{\emph{et al}\onedot}
\makeatother

\usepackage{graphicx}
\usepackage{booktabs}
\usepackage{microtype}
\usepackage{multirow}
\usepackage{bm}

\usepackage{xcolor}
\usepackage{soul}

\newcommand{\psp}{\kern0.2ex}
\newcommand{\nsp}{\kern-0.1ex}
\newcommand{\darrow}{$\downarrow$}
\newcommand{\uarrow}{$\uparrow$}
\newcommand{\tbf}[1]{\textls[-50]{\textbf{#1}}}
\newcommand{\tul}[1]{\underline{#1}}

\newcommand{\tworow}[2]{\begin{tabular}[c]{@{}c@{}}#1\vspace{-2pt}\\#2\end{tabular}}

\usepackage[accsupp]{axessibility}  % Improves PDF readability for those with disabilities.

\usepackage[pagebackref,breaklinks,colorlinks,citecolor=eccvblue]{hyperref}
\usepackage{orcidlink}

\usepackage{kotex} %%%%%% FOR KOREAN USE, WILL BE REMOVED

\begin{document}

\title{Kinetic Typography Diffusion Model} 

\titlerunning{Kinetic Typography Diffusion Model}

\author{Seonmi Park\orcidlink{0009-0007-8890-554X} \and
Inhwan Bae\kern0.05em\orcidlink{0000-0003-1884-2268} \and
Seunghyun Shin\orcidlink{0009-0006-3012-9675} \and
Hae-Gon Jeon\thanks{Corresponding author}\orcidlink{0000-0003-1105-1666}}

\authorrunning{S. Park et al.}

\institute{AI Graduate School, GIST, South Korea\\
\email{\{bluesky1000, inhwanbae, seunghyuns98\}@gm.gist.ac.kr, haegonj@gist.ac.kr}\\
\url{https://seonmip.github.io/kinety}
}

\maketitle

\begin{abstract}\
This paper introduces a method for realistic kinetic typography that generates user-preferred animatable ``text content''. We draw on recent advances in guided video diffusion models to achieve visually-pleasing text appearances. To do this, we first construct a kinetic typography dataset, comprising about 600K videos. Our dataset is made from a variety of combinations in 584 templates designed by professional motion graphics designers and involves changing each letter's position, glyph, and size (\ie, flying, glitches, chromatic aberration, reflecting effects, etc.). Next, we propose a video diffusion model for kinetic typography. For this, there are three requirements: aesthetic appearances, motion effects, and readable letters. This paper identifies the requirements. For this, we present static and dynamic captions used as spatial and temporal guidance of a video diffusion model, respectively. The static caption describes the overall appearance of the video, such as colors, texture and glyph which represent a shape of each letter. The dynamic caption accounts for the movements of letters and backgrounds. We add one more guidance with zero convolution to determine which text content should be visible in the video. We apply the zero convolution to the text content, and impose it on the diffusion model. Lastly, our glyph loss, only minimizing a difference between the predicted word and its ground-truth, is proposed to make the prediction letters readable. Experiments show that our model generates kinetic typography videos with legible and artistic letter motions based on text prompts.
\end{abstract}

\begin{figure}[t]
\centering
\includegraphics[width=\linewidth,trim={0 0 0 0},clip]{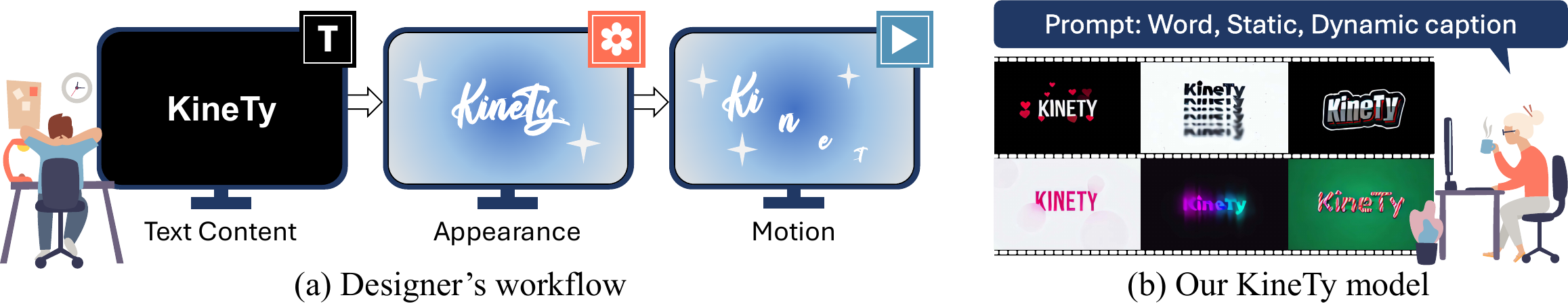}
\vspace{-5mm}
\caption{An overview of our KineTy pipeline, which is motivated by the designer's workflow. Our key idea is to generate eye-catching and aesthetic animatable words based on user instructions.
}
\label{fig:teaser}
\end{figure}

\section{Introduction}
\label{sec:introduction}
Kinetic typography is an artistic motion graphics design combining text and animations~\cite{ford1997kinetic}. Based on the word's meaning, motions are generated to convey information in videos.
The main goal is eye-catching and to improve message retention. With the rise of video media, it has become an essential element in TV programs, commercials, music videos and film leaders.

Kinetic typography controls letters' shape (glyph), color and texture over time, and transforms their positions. Professional motion graphic designers use commercial software like `Adobe After Effects' to make and render kinetic typography videos. The conventional pipeline is as follows~\cite{smith2012adobe,fridsma2019adobe}; (1) Define a text box in a workspace called composition and enter a text (\i.e. word or message). (2) Edit the static text by tuning its font, color and texture (3) Build the background if needed (4) Apply motion effects for all text and background (5) Repeat, optimize, and fine-tune the whole process until it satisfies user intention. This process is time-consuming and labor-intensive. It takes three hours for simple motions to several days for sophisticated effects per single kinetic typography video~\cite{kato2015textalive,fridsma2019adobe}. To efficiently do this, designers may need to consider either various design references or pre-defined design options called templates. 

With the advancement of generative models, there have been attempts to produce dynamic typography. The literature on typography generation has mostly focused on creating static single-letter images.
Starting from transforming the shape of the object into a single letter~\cite{iluz2023word,tanveer2023ds,wang2023anything,mu2024font},
there have been works to create multiple letters and multiple words by introducing a concept of layout~\cite{chen2023textdiffuser,chen2024textdiffuser,yang2023glyphcontrol}. Furthermore, DynTypo~\cite{men2019dyntypo} and Shape-Matching GAN\texttt{++}~\cite{yang2021shape} adopt style-transfer techniques to produce animatable effects on a single letter. They focus on only animatable effects without any actual motions of letters.

Meanwhile, the recent advancement of the video generation model has been actively studied with the success of text-guided video diffusion models, especially user's description conditions video frames~\cite{blattmann2023align,ge2023preserve,ho2022imagen,singer2022make,guo2023animatediff}. Although these models have opened up the possibility of creating kinetic typography, there is a critical issue that the video generations show a weak understanding of the letters' shapes and motions.

In this paper, we propose the \textbf{Kine}tic \textbf{Ty}pography diffusion model (called KineTy model) that generates kinetic typography from user-provided text prompts. Inspired by the video diffusion model, we allow users to input comprehensive descriptions for color, font, size, position and motion effects of letters. To better represent text motions, we first introduce our kinetic typography video dataset. 600K videos are rendered by combining randomly generated text contents with 584 templates made by professional motion graphic designers. These videos are labeled with static and dynamic captions that describe the video with respect to its appearance and motion characteristics, respectively. 

Next, we present KineTy model that effectively synthesizes videos from the text prompt. Here, we reevaluate how to effectively condition the caption guidance into the video diffusion model at a fundamental level. We enforce that each effect is incorporated by separately inserting static and dynamic captions into the spatial attention and temporal attention, respectively. To strengthen these attentions, we apply a zero convolution~\cite{zhang2023adding} into the word caption, and add it to them. We use a mask loss term only for the generated video by masking out the background contents, which makes the text contents more readable.

Experimental results shows that our KineTy robustly generates kinetic typography video with multi-letter legibility, while accurately representing captions. Furthermore, extensive and meticulous user studies support our claim that KineTy produces more aesthetic outcomes than general-purpose video generations.

\section{Related Work}
\label{sec:relatedworks}
\subsection{Typography Generation}
The key of typography is to make readable and visually-appealing text contents to readers. Most research in this field has focused on transferring the style of a single-letter design, including elements like color, glyph, font and effects, to other letters~\cite{tham2024vecfu, azadi2018multi,yang2017awesome,li2023compositional,nagata2023contour,liu2022learning,anderson2022neural,reddy2021multi,campbell2014learning,liu2021decoupled,pan2023few,chen2023joint,xia2023vecfontsdf,wang2021deepvecfont, wang2023deepvecfont,wang2023cf,liu2023dualvector,fu2023neural,berio2022strokestyles, li2021few, yang2023fontdiffuser}. Additionally, there have been works on transferring an image style to the text contents~\cite{yang2019controllable}. Similarly, research on scene text editing, which changes text content while maintaining its own style in a scene, is also underway~\cite{krishnan2023textstylebrush,qu2023exploring,yang2020swaptext,tuo2023anytext}, whose extended version to video is available in~\cite{subramanian2021strive}. In addition to changing text contents, color editing~\cite{shimoda2021rendering} and text segmentation~\cite{xu2021rethinking,wang2022self} are proposed.

With the recent success of the high-fidelity text-to-image diffusion model~\cite{rombach2022high}, typography generation has gained interest. Works in~\cite{tanveer2023ds, iluz2023word, wang2023anything} apply image styles into the target letter in an unsupervised manner.  

There are concurrent works that display multi-letters, beyond the single-letter generation. They follow the multi-step approach that firstly generates layouts at specific positions, arrange them with multi-letters or multi-words with the same fonts and colors, and contextually infers the background~\cite{chen2024textdiffuser,chen2023textdiffuser,shimoda2024towards,jia2023cole,zhang2023editing,wang2022aesthetic,xu2023unsupervised,jahanian2013recommendation, yang2023glyphcontrol}. However, these models are not specialized for typography. Since all letters are generated together, it is not editable for the letters, which makes it difficult to animate and add movement to each character.

\subsection{Typography Video Generation}
As we mentioned above, this work is the first attempt to generate kinetic typography. Although no existing works directly align with this, there are some works related to typography videos~\cite{minakuchi2005automatic,lee2006using,lee2002kinetic,wong1996temporal,kato2015textalive,xie2023creating}. Here, we would like to introduce two methods related to ours. 

DynTypo~\cite{men2019dyntypo} proposes a dynamic typography model that transfers the dynamic effect with realistic movements like fire and water on a specific uppercase English letter to others. Shape-Matching GAN\texttt{++}~\cite{yang2021shape} transfers an image style into a target letter by matching these shapes with structures of the target letter. However, the position and glyph of the letter are still fixed, with no actual movement between frames. Even under the condition of a single uppercase letter, representing various static and dynamic effects through text prompts remains a challenging issue.

By substituting the motion description with external user input, the task becomes more tractable. A work, Wakey-Wakey~\cite{xie2023wakey}, transfers the source  GIF (Graphics Interchange Format) motion to the target text. Here, graphic designers manually assign additional corresponding key points to achieve motion transfer. However, this process still requires direct human intervention.

\subsection{Text-to-Video Diffusion Models}
The video diffusion model produces a visually-plausible and photo-realistic video based on text conditioning~\cite{blattmann2023align, ho2022imagen, singer2022make, ge2023preserve, guo2023animatediff, wang2023lavie, esser2023structure, luo2023videofusion}. The main challenge in the video diffusion model is to maintain the temporal coherency between frames. The pioneer works for video diffusion is to add temporal blocks to the text-to-image model and to learn temporal coherency between video frames~\cite{ho2022imagen,singer2022make,ge2023preserve}. Another type of relevant studies use pretrained models to leverage the temporal block~\cite{guo2023animatediff, blattmann2023align, blattmann2023stable}. In particular, Guo \etal~\cite{guo2023animatediff} propose a motion module as the temporal block. It is designed as a plug-and-play module. After joint training with the pre-trained weight of stable diffusion~\cite{rombach2022high}, the module enables any text-to-image model to generate the video.

Existing typography has been mainly studied using static single letters. The similar work to kinetic typography is also single letters, with only variation of motion effect and almost no letter's movement exists. To initiate kinetic typography generation, we present our KineTy dataset and model that allows dynamic motion, effects and glyph deformation of multi-letters, which will be explained at the next section.

\begin{table}[t]
\caption{A summary of the kinetic typography datasets.}
\vspace{-3mm}
\centering\large
\resizebox{\textwidth}{!}{%
\begin{tabular}{@{~~~~}ccc@{~~~}c@{~~~}cc@{~~~}c@{~~}ccccc@{}c}
\toprule
Dataset & Domain & \#Samples & Video & \tworow{Static}{Caption} & \tworow{Dynamic}{Caption} & \tworow{Multi-}{letter} & \tworow{Text}{Appearance} & \!\!\!\!\!\tworow{Background}{~}\!\!\!\!\! & \tworow{Text}{Movements} & ~~ \\ \midrule
Multi-Content~\cite{azadi2018multi}         & Typography         & 10K  & -          & -          & -           & -          & \checkmark & -           & - \\
TextLogo3K~\cite{wang2022aesthetic}         & Typography         & 3K   & -          & -          & -           & -          & \checkmark & -           & - \\
TE141K~\cite{yang2020te141k}                & Typography         & 141K & -          & -          & -           & \checkmark & \checkmark & \checkmark  & - \\
TextSeg~\cite{xu2021rethinking}             & Text Segmentation  & 4K   & -          & -          & -           & \checkmark & \checkmark & \checkmark  & - \\
LAION-Glyph-10M~\cite{yang2023glyphcontrol} & Text-to-Image      & 10M  & -          & \checkmark & -           & \checkmark & \checkmark & \checkmark  & - \\
MARIO-10M~\cite{chen2024textdiffuser}       & Text-to-Image      & 10M  & -          & \checkmark & -           & \checkmark & \checkmark & \checkmark  & - \\
AnyWord-3M~\cite{tuo2023anytext}            & Text-to-Image      & 3M   & -          & \checkmark & -           & \checkmark & \checkmark & \checkmark  & - \\ 
CATER-GEN~\cite{hu2022make}                 & Image-to-Video      & 35K  & \checkmark & -          & \checkmark  & N/A        & N/A        & $\triangle$ & N/A \\
WebVid-10M~\cite{bain2021frozen}            & Text-to-Video      & 10M  & \checkmark & \checkmark & $\triangle$ & N/A        & N/A        & \checkmark  & N/A \\
Vimeo25M~\cite{wang2023lavie}               & Text-to-Video      & 25M  & \checkmark & \checkmark & $\triangle$ & N/A        & N/A        & \checkmark  & N/A \\
\midrule
\tbf{Ours}                                  & Kinetic Typography & 600K & \checkmark & \checkmark & \checkmark & \checkmark  & \checkmark & \checkmark  & \checkmark \\
\bottomrule
\end{tabular}
}
\vspace{-1mm}
\label{tab:dataset}
\end{table}

\section{Kinetic Typography Dataset}\label{sec:dataset}
We describe how to build the KineTy dataset. Unlike previous datasets that only cover single-lettered images~\cite{yang2020te141k}, our KineTy dataset has not only visual effects on multiple letters, but also their animations. We first provide detailed process to render kinetic typography videos using templates made by professional editors in~\cref{sec:dataset_rendering}. Next, we introduce an way to describe their appearance and motions through text prompts in~\cref{sec:dataset_captioning}. Lastly, we explain how to make ground-truth kinetic typography video for strictly fair comparisons in~\cref{sec:dataset_benchmark}. We summarize the difference between ours and existing datasets in~\cref{tab:dataset}.

\begin{figure}[t]
\centering
\includegraphics[width=\linewidth,trim={0 47mm 0 0},clip]{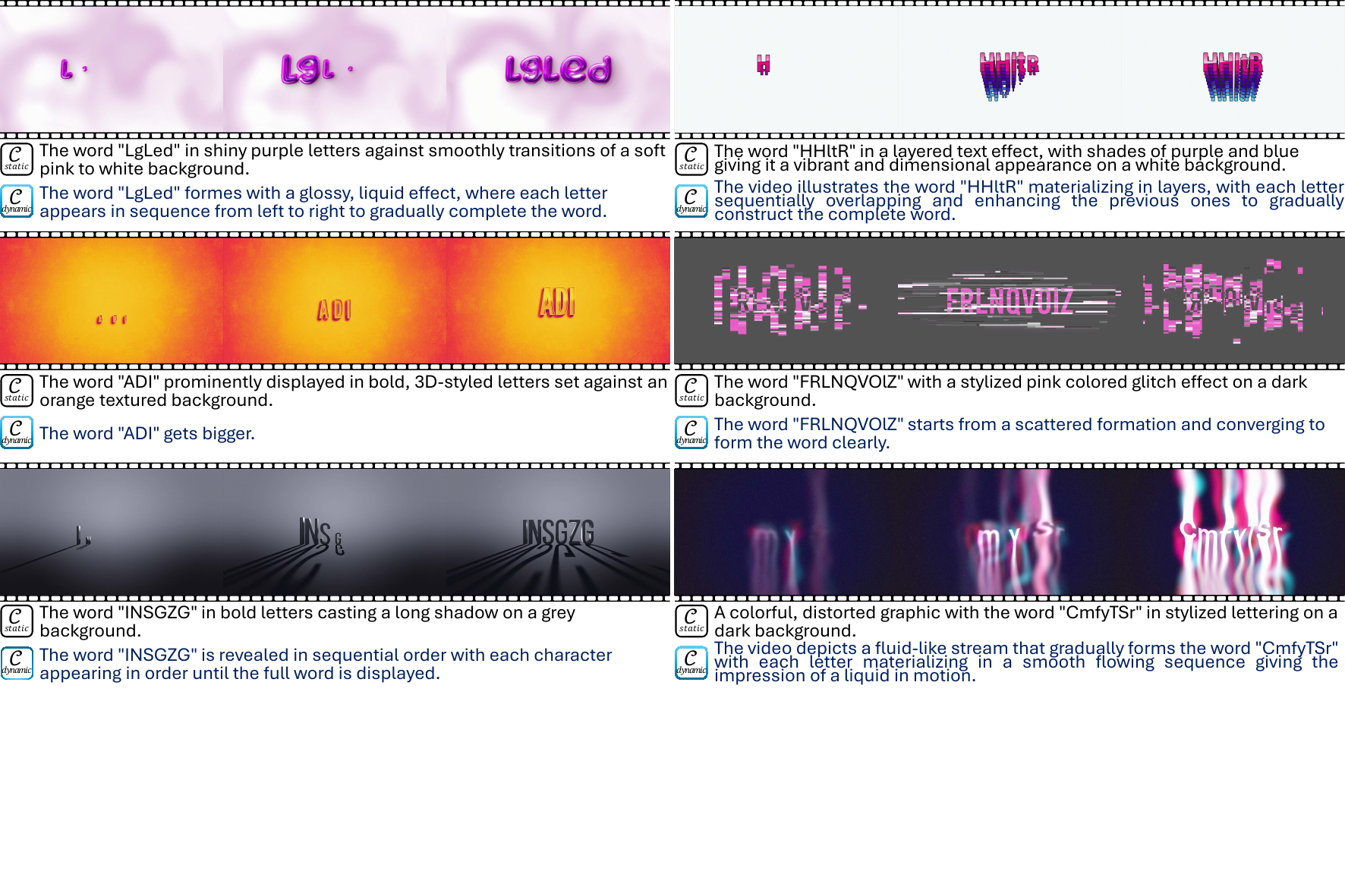}
\vspace{-5mm}
\caption{Examples of our KineTy dataset. Our dataset provides high-quality kinetic typography video created by professional motion graphic designers, along with captions that describe the visual appearance and motion effects. To aid visualization, we provide three frames from each video clip.}
\label{fig:dataset_example}
\end{figure}

\subsection{Video Rendering}\label{sec:dataset_rendering}
\noindent\textbf{Employing templates for video rendering.}
The template is a pre-designed project file which contains a visual effect on letters. Many editors prefer to use the templates because it saves their time and labor cost in practice. Following the best practice, we utilize 584 kinetic typography templates from professional graphic designers for our dataset construction, as visualized in~\cref{fig:dataset_example}.

\noindent\textbf{Set of multiple letters.}
Next, we utilize the kinetic typography templates and randomly replace the text contents for augmentation. We employ multiple letters, in contrast to existing typography datasets with static single-letters~\cite{azadi2018multi,yang2017awesome,li2023compositional}. Text contents are randomly generated by arranging up to 12 letters sampled from a set of 52 letters, including both uppercase and lowercase alphabets. Through this, we can expect rich letter-by-letter effects with various arrangements, while also keeping consistent styles across the multiple letters.

\noindent\textbf{Speculation.}
We render videos with 1,920$\times$1,080 resolution for 3 sec., and 1,000 random words from 584 templates. Subsequently, all videos are downsampled to 512$\times$288 resolution with 8fps for training. It takes a month to render the whole dataset with four i9 13900KF CPUs and an NVIDIA 2080ti GPU.

\begin{figure}[t]
\centering
\includegraphics[width=\linewidth,trim={0 129mm 38.5mm 0},clip]{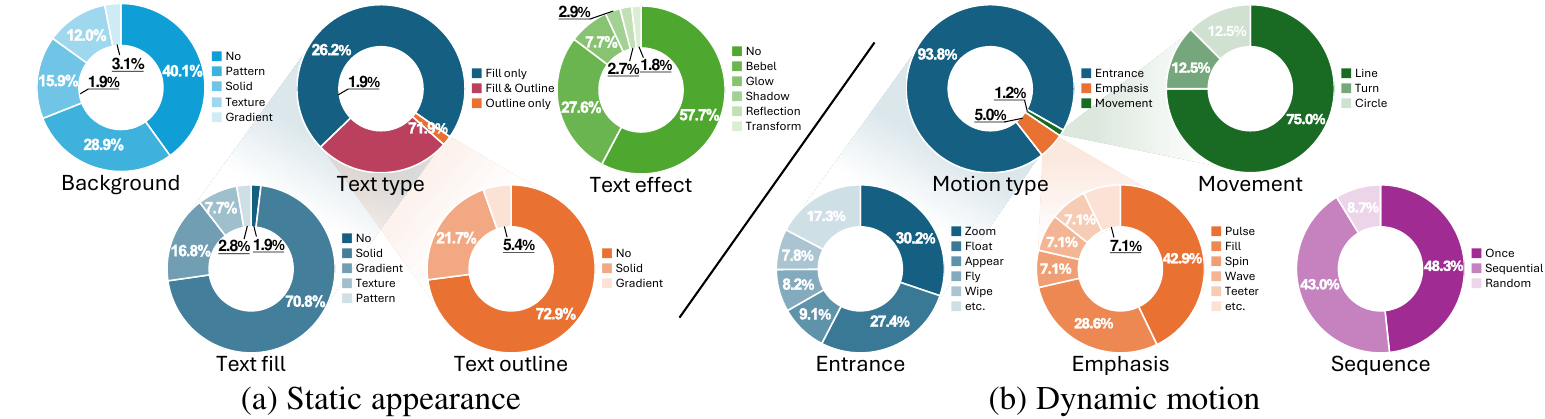}
\vspace{-6mm}
\caption{Statistics of our proposed dataset.}
\vspace{-2mm}
\label{fig:dataset_statistics}
\end{figure}

\subsection{Video Captioning.}\label{sec:dataset_captioning}
\noindent\textbf{Static and dynamic effect separation.}
In the next step, we caption the rendered kinetic typography for text-to-video generation. The professional editors usually design the static appearances at first, and then apply dynamic effects. Following this practice, we use two type of captions: static and dynamic captions, and labeled them accordingly. 

\noindent\textbf{Static effect captioning.}
To label static captions, we focus on letter's appearance which contains spatial information of typography.
As shown in~\cref{fig:dataset_example}, the static captions $\mathcal{C}_\textit{static}$ describe the color of the letters and the background, the glyph of the letters (e.g., outlined with yellow color or bold font), the characteristics of the background (e.g., textured and shiny background) and the arrangement of the letters (e.g., in a diagonal way) based on the last frame of the video where all the letters are displayed.

\noindent\textbf{Dynamic effect captioning.}
Similarly, we concentrate on the temporal change of motions for each frame to write dynamic captions. 
Dynamic captions $\mathcal{C}_\textit{dynamic}$ describe the motion part of the video, such as whether each letter appears in sequential or random order, can be rotated, or has a fade-in effect. 
For better systematic process, we initially label the videos with the GPT-4Vision model~\cite{gpt4vision}, and then manually verify and refine them to fix missing and wrongly labeled components.

\noindent\textbf{Statistics.}
The statistics of our dataset is summarized in \cref{fig:dataset_statistics}.
There are three main categories for the static appearance in \cref{fig:dataset_statistics} (a): Text type, text effect and background. 
Here, the text fill effects are dominant in the text type because it is the most obvious way to clearly show the text content. Especially, professional designers tend to prefer using solid colors, rather than text outlines.
The reason why 57.7\% of our dataset has no text effect is the readability of text contents. Since kinetic typography makes dynamic motions of the contents, users do not need for fancy visual effects on it.

In addition, we categorize dynamic motions into two-fold in \cref{fig:dataset_statistics} (b): motion type, including entrance, emphasis and movement, and sequence type: Our dataset supports various effect of entrance and emphasis to deliver striking messages. In contrast, the line has the majority in the movement because users' intention is usually conveyed after all letters in a scene are arranged in a line. 
In the same vein, when words come inside all at one and appear one-by-one in sequence, it is readable. They thus have the high portion in the sequence category.

\subsection{Ground-truth Video Generation}
\label{sec:dataset_benchmark}

Unlike generating random synthetic words in a training phase, we use real-world letters for evaluation by leveraging the templates. We can render ground-truth videos corresponding to user-input captions.

For evaluation, individual letters are selected for each alphabet letter from A to Z, and the first letters are capitalized for checking case sensitivity (\eg, Apple, Ball, ..., Zebra). These 26 words are used for rendering 584 templates, so that a total of 15,184 videos are used for the evaluation.

\begin{figure}[t]
\centering
\includegraphics[width=\linewidth,trim={0 0 0 0},clip]{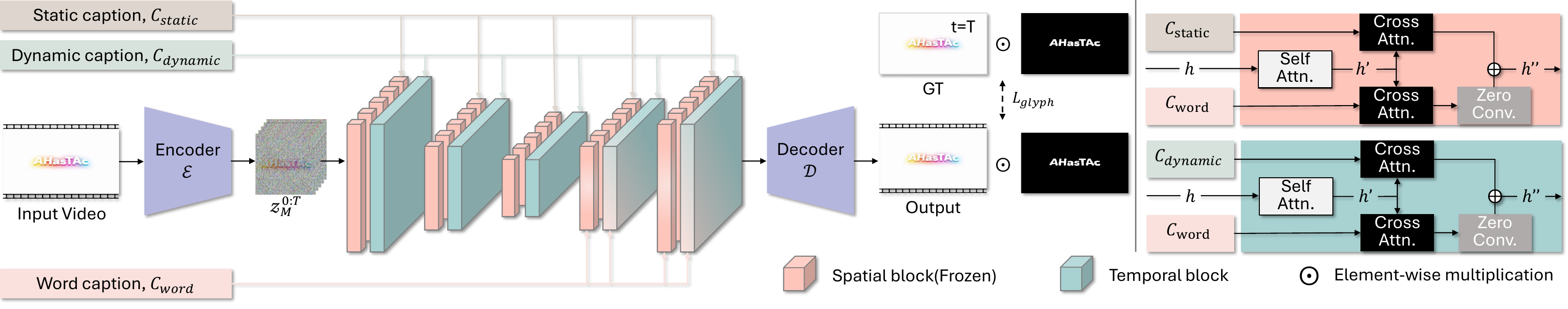}
\vspace{-6mm}
\caption{An architecture of our KineTy model.}
\label{fig:model}
\end{figure}

\section{Kinetic Typography Diffusion Model}
\label{sec:model}
In this section, we present our model for creating kinetic typography videos based on textual prompts. The key concept is to show animateble visual effects on texts, and our challenges in this work are summarized as: (1) the effect should be visually-pleasing and eye-catching to viewers; (2) the transition between frames should be smooth and align with the captions; (3) the text must be readable.

We begin by defining a kinetic typography generation problem in~\cref{sec:model_preliminary}. We then describe how to model static and dynamic captions that efficiently guide appearance and motions in~\cref{sec:model_caption}. We lastly discuss how to improve the glyph legibility with respect to the model design and its learning strategy in~\cref{sec:model_enhance}. Implementation details are provided in~\cref{sec:model_details}.

\subsection{Preliminary}\label{sec:model_preliminary}
\noindent\textbf{Conditional Latent Diffusion Models.}
Diffusion models train the data distribution by progressively refining a noisy initial state $z_M\sim \mathcal{N}(0,1)$ into the target data representation $z_0$ for $M$ diffusion steps. Recent advancements, particularly in Latent Diffusion Models (LDMs)~\cite{rombach2022high}, enhance the efficiency by encoding an image $x$ into a compact latent representation $z_0=\mathcal{E}(x)$ using an encoder $\mathcal{E}$, and is transformed back to the image $\tilde{x}=\mathcal{D}(z_0)$ using a decoder $\mathcal{D}$. These adding noise and its subsequent removal are done with U-Net-style denoising network $\epsilon_\pi$. In addition, a condition $y$ is mapped to the hidden state $h$ of $\epsilon_\pi$ via attention:
\begin{equation}
\begin{gathered}
    \text{Attn}_\theta(h,y) = \text{Softmax}\bigg(\frac{Q_\theta(h)K_\theta(y)^{\!\top}}{\sqrt{d}}\bigg)\cdot V_\theta(y) \\
    \text{s.t.} ~~~ Q_\theta(h) = W_{\!\theta\nsp,Q} \cdot h,~ K_\theta(y)=W_{\!\theta\nsp,K} \cdot y,~ V_\theta(y) = W_{\!\theta\nsp,V} \cdot y,
\end{gathered}
\label{eq:attention}
\end{equation}
where $W_{\!\theta\nsp,Q}$, $W_{\!\theta\nsp,K}$ and $W_{\!\theta\nsp,V}$ denote the learnable parameters for queries, keys and values. $d$ is the number of dimensions of the keys.
The objective function for the denoising network $\epsilon_\pi$ is formulated as:
\begin{equation}
    L_\textit{ldm} = \mathbb{E}_{\mathcal{E}(x),y,\epsilon \sim \mathcal{N}(0,1),m}
    \left[ \big\lVert
          \epsilon\,-\,\epsilon_\pi(z_m, m, y)
    \big\rVert_2^2\right],
\end{equation}
where $\|\cdot\|_2$ is the $L_2$ distance and diffusion step $m\in\{1, ..., M\}$ is uniformly sampled during training.

\noindent\textbf{Extending image diffusion models to video.}
Extending the capability of text-to-image diffusion models to video generation has been more feasible. By incorporating a temporal self-attention called motion module~\cite{guo2023animatediff}, it learns a temporal coherency between $T$ frames in a latent sequence $\bm{\mathrm{z}}_0^{0:T}$, which corresponds to the video sequence $\tilde{\bm{\mathrm{x}}}^{0:T}=\mathcal{D}(\bm{\mathrm{z}}_0^{0:T})$. This allows diffusion models to generate smoothly changed sequential images over time, whose temporal self-attention can be defined as follows:
\begin{equation}
    \text{Attn}_\phi(\bm{\mathrm{h}}^{0:T},\bm{\mathrm{h}}^{0:T}) = \text{Softmax}\bigg(\frac{Q_\phi(\bm{\mathrm{h}}^{0:T})K_\phi(\bm{\mathrm{h}}^{0:T})^{\!\top}}{\sqrt{d}}\bigg)\cdot V_\phi(\bm{\mathrm{h}}^{0:T}).
\label{eq:motion_module}
\end{equation}

\subsection{Spatial and Temporal Guidance}\label{sec:model_caption}
\noindent\textbf{Static caption incorporation.}
Motivated by the best practice of professional designers who handle appearance and motion effects separately, we divide the caption into static and dynamic elements.

When existing text-to-image models learn the distribution of image data, they are conditioned on texts related to the image's appearance. In the same manner, our model is also guided by captions describing the appearance of the text contents and background in each frame of video. Using~\cref{eq:attention}, we define a spatial attention block with a self-attention followed by a cross-attention with a static caption $\mathcal{C}_\textit{static}$ as: 
\begin{equation}
    \bm{\mathrm{h}}\nsp'^{\,0:T} = \big\{\text{Attn}_\vartheta\big(
    \text{Attn}_\theta(\bm{\mathrm{h}}^{t},\bm{\mathrm{h}}^{t}), \tau(\mathcal{C}_\textit{static})
    \big)\big\}_{t\psp=\psp0}^{\!T\nsp-1},
\label{eq:static_caption_attention}
\end{equation}
where $\tau(\cdot)$ is the CLIP text encoder~\cite{radford2021learning}.

\noindent\textbf{Dynamic caption incorporation.}
A previous work~\cite{guo2023animatediff} uses a motion module that only utilizes a self-attention between frames to learn the temporal consistency. On the other hand, in kinetic typography, it is essential to accurately display the dynamic motion effects of each letter in a video, following user's textual description. To do this, we extend~\cref{eq:motion_module} by adding the cross-attention with a dynamic caption as follows:
\begin{equation}
    \bm{\mathrm{h}}\nsp'^{\,0:T} = \text{Attn}_\psi\big(
    \text{Attn}_\phi(\bm{\mathrm{h}}^{0:T},\bm{\mathrm{h}}^{0:T}), \tau(\mathcal{C}_\textit{dynamic})
    \big),
\label{eq:dynamic_caption_attention}
\end{equation}
Through this process, the model becomes capable of maintaining temporal consistency as well as direct control over dynamic movements.

\subsection{Enhancing Glyph Legibility.}\label{sec:model_enhance}
\noindent\textbf{Readability improvement.}
As a next step, we aim to make the legibility of our model better. To do this, we first classify descriptions for text contents and prompt because diffusion models often have a difficulty in distinguishing them. In this work, we put down a delimiter between each character for text contents.

Let a text content $\mathcal{L} = \{ l_1,  l_2, ..., l_L\}$ contain $L$ letters. To condition each letter separately, we join each letter $l_i$ in the text content with a delimiter symbol `$|$'. Further, we add a symbol `\^{}' behind upper case letters because the clip encoder does not care of upper cases. To the end, when the word $\mathcal{L}$ denotes ``Apple'', $\mathcal{L}'$ is represented as ``A\!\^{}|p|p|l|e''. We denote the whole $\mathcal{L}$ as $\mathcal{L}'$ for better representation.

\noindent\textbf{Attention on text contents.}
It can be challenging for the attention module to guide both text contents and effects at once, while disentangling the text content information in CLIP feature space. What is worse, the text content can be overwhelmed by relatively long appearance and motion effects in CLIP text encoder. To tackle this problem, we introduce an additional cross-attention branch to the text content. Inspired by ControlNet~\cite{zhang2023adding}, we regard the text content as conditions through the zero convolution operation. Starting with the definition of word caption $\mathcal{C}_\textit{word}$ via a prompt template like ``The word $\{\mathcal{L}'\}$'', we extend~\cref{eq:static_caption_attention} by adding the cross-attention module between the word caption and hidden feature, weighted by the zero-initialized convolutions $\rho$ as follows:
\begin{equation}
\begin{gathered}
    \bm{\mathrm{h}}\nsp''^{\,0:T}_\textit{static} = \Big\{\text{Attn}_\vartheta\big(
        \bm{\mathrm{h}}\nsp', \tau(\mathcal{C}_\textit{static})
    \big) + 
    \rho_\upsilon\Big(\text{Attn}_\upsilon\big(
    \bm{\mathrm{h}}\nsp', \tau(\mathcal{C}_\textit{word})
    \big)\Big)
    \Big\}_{t\psp=\psp0}^{T\nsp-1}, \\\vspace{-2mm}
    \text{where} ~~~ \bm{\mathrm{h}}\nsp' = \text{Attn}_\theta(\bm{\mathrm{h}}^{t},\bm{\mathrm{h}}^{t}).
\end{gathered}
\end{equation}
This allows the network to gradually evaluate the usefulness of this additional caption based on the condition. In the same way, we incorporate the word caption into the temporal cross-attention in~\cref{eq:dynamic_caption_attention} as:
\begin{equation}
\begin{gathered}
    \bm{\mathrm{h}}\nsp''^{\,0:T}_\textit{dynamic} = \text{Attn}_\psi\big(
        \bm{\mathrm{h}}\nsp', \tau(\mathcal{C}_\textit{dynamic})
    \big) + 
    \rho_\varphi\Big(\text{Attn}_\varphi\big(
    \bm{\mathrm{h}}\nsp', \tau(\mathcal{C}_\textit{word})
    \big)\Big), \\\vspace{-2mm}
    \text{where} ~~~ \bm{\mathrm{h}}\nsp' = \text{Attn}_\phi(\bm{\mathrm{h}}^{0:T},\bm{\mathrm{h}}^{0:T}).
\end{gathered}
\end{equation}

\noindent\textbf{Glyph loss.}
To train our model, we use a common loss function $L_\textit{ldm}$. Additionally, we impose an extra penalty on the letter regions to enforce a sharp and correct glyph of text contents. We first use a binary mask $B$ for the letters in the last frame $V^T$ from a text segmentation model~\cite{xu2021rethinking}. The mask is then blurred to cover its surrounding effects. We then define a glyph loss based on $L_{LDM}$ with the additional pixel-wise weighting strategy using the blurred mask as follows:
\begin{equation}
    L_\textit{glyph} = \mathbb{E}_{\mathcal{E}(x),y,\epsilon\sim\mathcal{N}(0,1),m} \left[ \big\lVert B\odot\big(\epsilon\,-\,\epsilon_\pi(z_m^T, m, y)\big) \big\rVert_2^2\right],
\end{equation}
where $\odot$ is the element-wise multiplication.

With the glyph loss, we enhance the legibility of text contents, enabling the precise creation of multi-letter formations. Through the linear combination of both loss terms, we can formulate the final loss function as $L = L_\textit{ldm} + \alpha L_\textit{glyph}$. Here, we empirically set $\alpha$ to 0.01.

\subsection{Implementation Details}\label{sec:model_details}
To train our model, we use a two-step training strategy similar to that of human trainees~\cite{fridsma2019adobe}. First, we pre-train the network to generate spatial appearances using only static caption and last video frame pairs. Here, we detach the temporal attention modules to make them work as text-to-image diffusion models during 30 epochs with a batch size of 200. Next, we train the full model while freezing the spatial attention using whole video frames and captions for an epoch with a batch size of 8. For training, we resize the height $H$ and width $W$ of the video into 256$\times$256 with $T=24$ frames. The training is performed with AdamW optimizer~\cite{loshchilov2018decoupled}, with a diffusion step $M$ of 1000 and a learning rate of 0.0001, which usually takes about 20 hours for training on 8 NVIDIA A100 GPUs. 
The inference time is about 20 seconds when the number of sampling steps is 25.

\section{Experiments}\label{sec:experiment}
In this section, we conduct comprehensive experiments to verify the effectiveness of our model for kinetic typography. We first describe the experimental setup in~\cref{sec:experiment_setup}. We then provide comparison results with relevant typography and video generation models, and report user-study to highlight the practical applicability of our model in~\cref{sec:experiment_result}. We lastly carry out an extensive ablation study to validate the effect of each component in our model in~\cref{sec:experiment_ablation}

\subsection{Experimental Setup}\label{sec:experiment_setup}
\noindent\textbf{Comparison methods.}
We compare ours with state-of-the-art generative models, which consist of 3 two-stage methods and 2 one-stage methods. The two-stage methods are based on combinations of text-to-image models, including DS-Fusion~\cite{tanveer2023ds}, GlyphControl~\cite{yang2023glyphcontrol} and Text-Diffuser~\cite{chen2024textdiffuser}, and an image-to-video model,~SparseCtrl~\cite{guo2023sparsectrl}. DS-Fusion outputs a stylized letter based on the given style phrase and a letter. Since our caption is a sentence, we extract a style keyword from it and consider it as a style phrase. Note that DS-Fusion is able to generate one letter at a time, so we concatenate each letter to make a full-text content. GlyphControl takes an instruction to generate a glyph image for text contents, then uses the image as a condition for the final output image. Text-Diffuser first finds a proper layout for words and produce the output based on a caption and the layout. Since the comparison methods yield images, we combine them with an image-to-video model, SparseCtrl, to compare with ours. SparseCtrl uses a caption and a guidance such as sketches, depth maps and images to generate a stylized video. 

On the other hand, one-stage-methods, AnimateDiff~\cite{guo2023animatediff} and Lavie~\cite{wang2023lavie}, are based on text-to-video models. AnimateDiff extends pre-trained text-to-image diffusion models with a motion module, and trains only the motion module to make fully use of the massive information of video dataset. Lavie leverages a temporal self-attention block to enhance a temporal consistency between frames while keeping the generation performance.

For a fair comparison, we utilize the official source codes and pre-trained weights provided by the authors. For more analysis, we categorize the motion effects into three groups, `Entrance', `Emphasis' and `Motion', and conduct experiments to check the quality of each result.

\newcommand{\FVD}{\nsp\textls[-10]{FVD}\!\,\raisebox{0.15ex}{\darrow}\!\nsp}
\newcommand{\IS}{\nsp\kern 0.61em IS\raisebox{0.15ex}{\darrow}\!\,\kern 0.61em\!\nsp}
\newcommand{\CLIP}{\nsp\textls[-30]{CLIP}\!\,\hspace{-0.1em}\raisebox{0.15ex}{\uarrow}\!\nsp}
\newcommand{\OCR}{\nsp\textls[-20]{OCR}\!\,\hspace{-0.02em}\raisebox{0.15ex}{\uarrow}\!\nsp}
\newcommand{\GAI}{\kern0.5ex Gain\raisebox{0.15ex}{\uarrow}\kern0.5ex}

\newcommand{\dsfpsvd}{DS-F\!\!\cite{tanveer2023ds}\texttt{+}Stable\!\!\cite{blattmann2023stable}}
\newcommand{\glypsvd}{Glyph\!\!\cite{yang2023glyphcontrol}\texttt{+}Stable\!\!\cite{blattmann2023stable}}
\newcommand{\tdfpsvd}{T-Diff\!\!\cite{chen2024textdiffuser}\texttt{+}Stable\!\!\cite{blattmann2023stable}}
\newcommand{\dsfpsps}{DS-F\!\!\cite{tanveer2023ds}\texttt{+}Sparse\!\!\cite{guo2023sparsectrl}}
\newcommand{\glypsps}{Glyph\!\!\cite{yang2023glyphcontrol}\texttt{+}Sparse\!\!\cite{guo2023sparsectrl}}
\newcommand{\tdfpsps}{T-Diff\!\!\cite{chen2024textdiffuser}\texttt{+}Sparse\!\!\cite{guo2023sparsectrl}}
\newcommand{\animdif}{AnimateDiff\!\!\cite{guo2023animatediff}}
\newcommand{\lavie}{Lavie\!\cite{wang2023lavie}}

\begin{table}[t]
\caption{Quantitative evaluation of two-stage generation methods, and 
one-stage text-to-video generation methods. \tbf{Bold}: Best, \tul{Underline}: Second-best.}
\vspace{-3mm}
\centering\large
\resizebox{\linewidth}{!}{
\begin{tabular}{c|c cccc c cccc c cccc c|c cccc }
\toprule
\multirow{2}{*}{Model\vspace{-5pt}} & & \multicolumn{4}{c}{Entrance} & ~ & \multicolumn{4}{c}{Emphasis} & ~ & \multicolumn{4}{c}{Motion} & & & \multicolumn{4}{c}{Average} \\ \cmidrule{3-6} \cmidrule{8-11} \cmidrule{13-16} \cmidrule{19-22}
              & & \FVD& \IS &\CLIP& \OCR& & \FVD& \IS &\CLIP& \OCR& & \FVD& \IS &\CLIP& \OCR& & & \FVD& \IS &\CLIP& \OCR \\ \midrule
\dsfpsps      & & 1636.3 & 5.71 & 0.69 & 0.58 & & 3184.9 & 5.97 & 0.73 & 0.58 & & 1236.5 & 5.42 & 0.76 & 0.56 & & & 1843.10 & 5.66 & 0.70 & 0.58 \\ 
\glypsps      & & 2415.6 & 4.30 & 0.76 & 0.68 & & 2667.1 & 5.06 & 0.76 & \tul{0.75} & & 1468.6 & 4.48 & 0.84 & 0.80 & & & 2361.98 & 4.46 & 0.76 & 0.69 \\
\tdfpsps      & & 2937.2 & 5.61 & 0.78 & \tbf{0.90} & & 3107.5 & 5.16 & 0.79 & \tbf{0.89} & & 1208.2 & 4.06 & 0.82 & \tul{0.95} & & & 2815.6 & 5.35 & 0.78 & \tbf{0.90} \\ \cmidrule(lr){1-22}
\animdif      & & 1613.8 & 4.71 & 0.59 & 0.01 & & 2106.0 & 4.56 & 0.66 & 0.03 & & 2160.6 & 4.51 & 0.67 & 0.07 & & & 1727.55 & 4.68 & 0.61 & 0.01 \\
\lavie        & & \tul{825.1} & \tul{2.25} & \tul{0.82} & 0.54 & & \tul{1835.4} & \tul{3.83} & \tul{0.83} & 0.41 & & \tul{505.44} & \tul{3.33} & \tul{0.85} & 0.35 & & & \tul{956.97} & \tul{2.68} & \tul{0.82} & 0.49 \\ \midrule
\textbf{Ours} & & \tbf{147.4} & \tbf{1.79} & \tbf{0.87} & \tul{0.71} & & \tbf{177.52} & \tbf{1.75} & \tbf{0.88} & 0.54 & & \tbf{36.16} & \tbf{1.62} & \tbf{0.94} & \textbf{0.98} & & & \tbf{125.90} & \tbf{1.77} & \tbf{0.88} & \tul{0.76} \\ \bottomrule
\end{tabular}
}
\vspace{-2mm}
\label{tab:gpgraph_trajresult}
\end{table}

\begin{figure}[t]
\centering
\includegraphics[width=\linewidth,trim={0 0 0 0},clip]{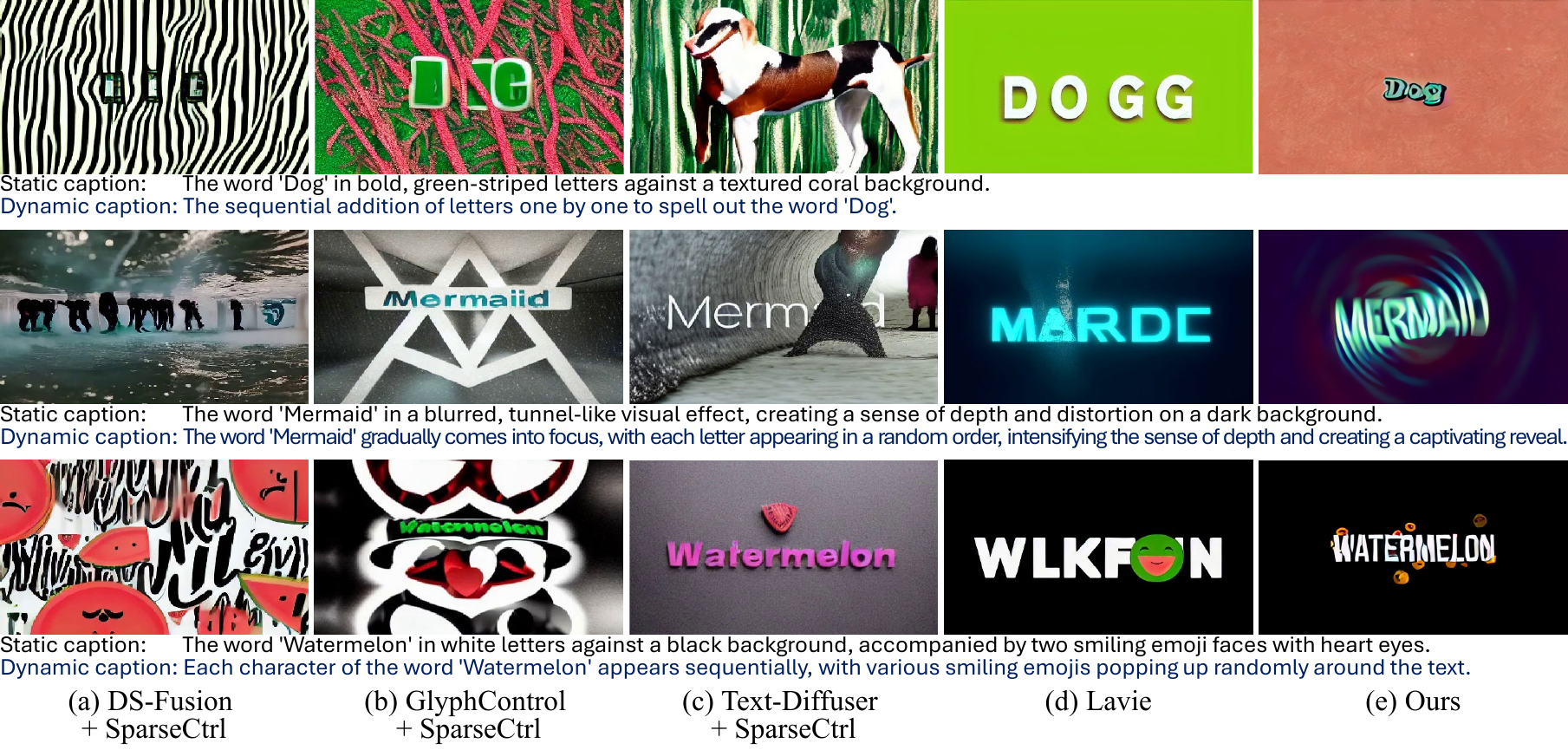}
\vspace{-7mm}
\caption{Qualitative results from the comparison models and ours. Obviously, the results from ours reflect the contents of captions better than the others.}
\vspace{-5mm}
\label{fig:experiment_comparison}
\end{figure}

\noindent\textbf{Evaluation metrics.}
Since the final products of kinetic typography depend on the designer's preferences, we adopt four feature-based metrics rather than photo-consistency measures like PSNR. (1) Fr\'echet Video Distance (FVD) qualifies a similarity between two videos by comparing their feature representations extracted from a pre-trained deep neural network~\cite{unterthiner2018towards}. (2) Inception Score (IS) evaluates the quality and diversity of images generated by a model~\cite{barratt2018note}. (3) CLIPScore measures the semantic alignment between generated images and their textual descriptions, using the CLIP model~\cite{radford2021learning, huang2021unifying, hessel2021clipscore}. (4) Optical Character Recognition (OCR) checks the clarity of the output by comparing recognized text to a reference text using the F1-score.

\subsection{Evaluation Results}\label{sec:experiment_result}
\begin{figure}[t]
\centering
\includegraphics[width=\linewidth,trim={0 0 0 0},clip]{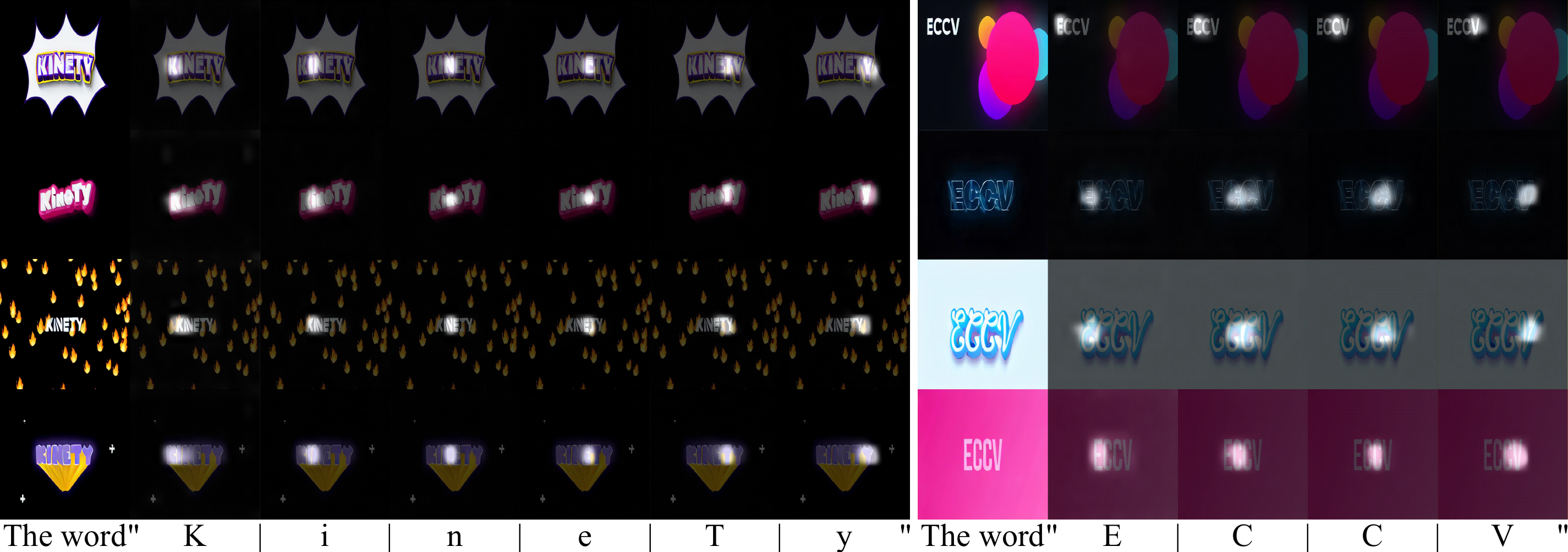}
\vspace{-6mm}
\caption{Visualization of attention maps between letters from a caption and feature map of a hidden layer.}
\vspace{-4mm}
\label{fig:experiment_attention}
\end{figure}

\noindent\textbf{Quantitative results.}
As shown in~\cref{tab:gpgraph_trajresult}, our KineTy model achieves the better performance than the comparison methods in the most metrics on all the categories. The comparison methods, despite their impressive performance in general image and video generation, show unsatisfactory performance in creating kinetic typography. The reasons are mainly two-folds:
(1) Text-to-video models are not specialized for kinetic typography and often fail to exactly distinguish the text content in the long captions. Different from them, we additionally feed $\mathcal{C}_{word}$ which imposes a high attention to the text content in the video. (2) They have challenges in creating letter-level motions. Since they are unaware of each letter in the text content, all the letters move as a whole or inconsistently. We handle this issue by providing a delimiter between each letter and using zero-convolution in~\cref{fig:experiment_attention}.

Our model shows promising results in generating legible text content, but it sometimes falls short in OCR accuracy. This is because the two-stage methods algorithmically generate a glyph image of a text content, and then deform glyphs conditioned on the glyph image. As a result, they have the better performance in OCR metrics than ours and one-stage methods. Nevertheless, our method outperforms in the most cases, demonstrating the effectiveness of our method in generating the clear text content and motion effects.

\begin{figure}[t]
\centering
\includegraphics[width=\linewidth,trim={0 0mm 0 0},clip]{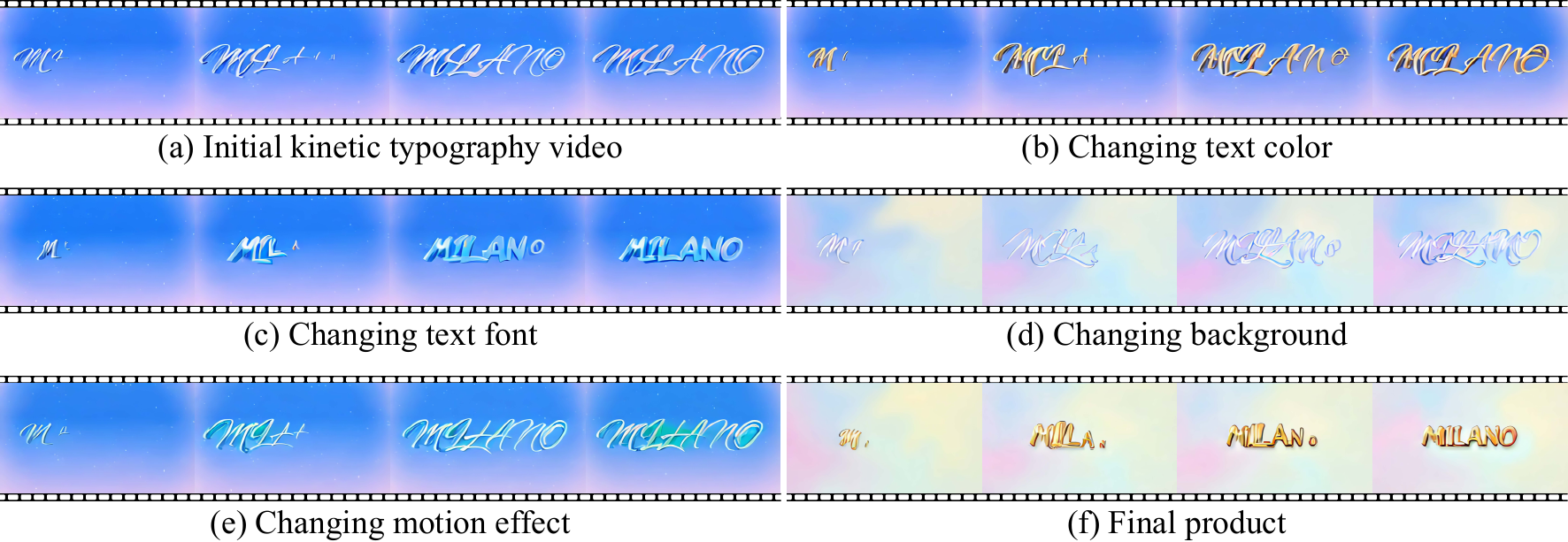}
\vspace{-6mm}
\caption{The demonstration of the editable capability of our KineTy model after generating initial kinetic typography.}
\vspace{-6mm}
\label{fig:generality}
\end{figure}

\noindent\textbf{Qualitative results.}
In~\cref{fig:experiment_comparison}, we display three examples of the comparison methods. Compared to ours, the two-stage methods do not understand the user's instruction with respect to either the whole caption or the letter's movement. Here, we need to know how to perform our KineTy model. \Cref{fig:experiment_attention} visualizes the attention map between the caption and features from the hidden state on the corresponding images. Thanks to the cross-attention between the word caption and the noisy latent, and the deliminator to separate the words, our model successfully embeds features for each letter in the noise.

Lastly, to show the generality of our KineTy model, we conduct an additional experiment. After generating an initial kinetic typography, our KineTy model allows users to modify its style and motions such as the content color, background, font and motion, and even to combine them. As demonstrated in~\cref{fig:generality}, ours can render the modified outcomes corresponding to the additional captions.

\begin{table}[t]
\caption{The result of user-study. The score ranges from 1 $\sim$ 6 for Study 1 \& 2, and 0 $\sim$ 1 for Study 3. \tbf{Bold}: Best, \tul{Underline}: Second-best.}
\vspace{-3mm}
\centering\large
\resizebox{\linewidth}{!}{
\begin{tabular}{c|c|c cccccc c cccccc}
\toprule
\multirow{2}{*}{~~~~\tworow{User}{Study}~~~~\vspace{-17pt}} & \multirow{2}{*}{~~~~\tworow{Dynamic}{Effect}~~~~\vspace{-17pt}} & ~ & \multicolumn{6}{c}{General} & ~~~ & \multicolumn{6}{c}{Expert} \\ 
\cmidrule{4-9} \cmidrule{11-16} 
         & & & \tworow{DS-F}{\textls[-10]{\texttt{+}Sparse}} & \tworow{Glyph}{\textls[-10]{\texttt{+}Sparse}} & \tworow{T-Diff}{\textls[-10]{\texttt{+}Sparse}} & \tworow{\textls[-20]{\!A\nsp nimate\!}}{Diff} & ~Lavie~ & ~~~\textbf{Ours}~ & & \tworow{DS-F}{\textls[-10]{\texttt{+}Sparse}} & \tworow{Glyph}{\textls[-10]{\texttt{+}Sparse}} & \tworow{T-Diff}{\textls[-10]{\texttt{+}Sparse}} & \tworow{\textls[-20]{\!A\nsp nimate\!}}{Diff} & ~Lavie~ & ~~~\textbf{Ours}~~ \\ \midrule
\multirow{4}{*}{Study\,1\vspace{-5pt}} & 
    Entrance & & 3.384 & 3.687 & \tul{3.891} & 2.406 & 3.153 & ~~\tbf{4.478}  & & 3.481 & 4.025 & \tul{4.131} & 1.019 & 3.081  & ~~\tbf{5.263}~ \\
 & Emphasis & & 3.562 & 3.713 & \tul{4.119} & 2.563 & 2.725 & ~~\tbf{4.319}& & 3.625 & 4.175 & \tul{4.225} & 1.050 &2.625  & ~~\tbf{5.300}~ \\
 & Motion   & & 2.363 & \tul{4.138} & 4.113 & 3.338 & 2.375 & ~~\tbf{4.675} & & 1.700 & 3.550 & 4.000 & 4.000 & 2.000 & ~~\tbf{5.750}~ \\ \cmidrule(lr){2-16}
 & Average  & & 3.318 & 3.713 & \tul{3.923} & 2.518 & 3.036 & ~~\tbf{4.494} & & 3.335 & 3.995 & \tul{4.090} & 1.325 & 2.905 & ~~\tbf{5.350}~ \\ \midrule
\multirow{4}{*}{Study\,2\vspace{-5pt}} & 
 Entrance & & 3.514 & 3.775 & \tul{3.941} & 2.413 & 3.270 & ~~\tbf{4.088} & & 3.650 & 4.081 & \tul{4.244} & 1.019 & 3.106 & ~~\tbf{4.900}~ \\
 & Emphasis & & 3.531 & 3.881 & \tul{4.075} & 2.269 & 3.050 & ~~\tbf{4.194} & & 3.725 & \tul{4.150} & 4.050 & 1.025 & 2.675 & ~~\tbf{5.375}~ \\
 & Motion   & & 2.838 & \tul{3.950} & 3.875 & 3.250 & 2.850 & ~~\tbf{4.238} & & 2.150 & 4.000 & 4.450 & 3.450 & 1.950 & ~~\tbf{5.000}~ \\ \cmidrule(lr){2-16}
 & Average  & & 3.439 & 3.799 & \tul{3.948} & 2.460 & 3.220 & ~~\tbf{4.135}  & & 3.485 & 4.060 & \tul{4.255} & 1.265 & 2.970 & ~~\tbf{4.995}~ \\  \midrule
\multirow{4}{*}{Study\,3\vspace{-5pt}} & 
 Entrance & & 0.628 & 0.620 & \tul{0.652} & 0.063 & 0.169 & ~~\tbf{0.725} & & 0.844 & 0.856 & \tul{0.887} & 0.019 & 0.138 & ~~\tbf{0.969}~ \\
 & Emphasis & & 0.581 & 0.606 & \tul{0.638} & 0.088 & 0.106 & ~~\tbf{0.781} & & 0.750 & 0.700 & \tul{0.900} & 0.000 & 0.000 & ~~\tbf{1.000}~ \\
 & Motion   & & 0.075 & \tul{0.713} & 0.750 & 0.362 & 0.288 & ~~\tbf{0.781}& & 0.000 & \tul{0.900} & \tul{0.900} & 0.500 & 0.250 & ~~\tbf{1.000}~ \\ \cmidrule(lr){2-16}
 & Average  & & 0.553 & 0.626 & \tul{0.658} & 0.096 & 0.174 & ~~\tbf{0.744} & & 0.730 & 0.835 & \tul{0.880} & 0.065 & 0.135 & ~~\tbf{0.975}~ \\ 
\bottomrule
\end{tabular}
}
\vspace{-6mm}
\label{tab:user_study1}
\end{table}

\noindent\textbf{User Study.}
To assess the practical utility of our results in the field of kinetic typography, we conduct a user study using Amazon MTurk. We ask 20 questionnaires to 50 participants. Since their familiarity of typography can vary significantly, we divide them into two groups: experts and non-experts, consisting of 10 and 40 individuals, respectively. The study involves three sections: (1) caption alignment; (2) kinetic typography suitability; (3) Word readability. We make two video clips by randomly choosing two words for each effect. In total, we make 20 clips for this user study.

In the first study, we measure how well ours and the comparison methods generate videos that align with provided captions. The participants see 6 videos from different models with the same corresponding caption in a random order. They are asked to rank the videos from 1 to 6, and the rankings are subsequently converted into scores, ranging from 6 (Best alignment) to 1 (Worst alignment). 
The following study evaluates how proper the generated videos are for motion graphic applications. Since some participants might be unfamiliar with kinetic typography, all participants watch $4$ example videos from online in advance. After that, they check six videos in a random order and rank them. Scores were assigned based on these rankings, from 6 (the most proper) to 1 (the least proper).
The last study assesses how readable the outcomes are. After watching 6 videos, the participants are needed to vote for videos with readable text contents. Of course, multiple voting is available.

The performance of each method is measured based on scores for the first two studies, and on the selection ratio for third one. As demonstrated in~\cref{tab:user_study1}, our approach shows the promising performance across all studies. The results highlight our model's practical usability in creating kinetic typography, especially for the domain experts who give our model the highest scores.

\begin{table}[t]
\caption{The result of ablation study. \tbf{Bold}: Best, \tul{Underline}: Second-best.}
\vspace{-3mm}
\centering\large
\resizebox{\linewidth}{!}{
\begin{tabular}{c|c cccc c cccc c cccc c|c cccc }
\toprule
\multirow{2}{*}{Model\vspace{-5pt}} & & \multicolumn{4}{c}{Entrance} & ~ & \multicolumn{4}{c}{Emphasis} & ~ & \multicolumn{4}{c}{Motion} & & & \multicolumn{4}{c}{Average} \\ \cmidrule{3-6} \cmidrule{8-11} \cmidrule{13-16} \cmidrule{19-22}
              & & \FVD& \IS &\CLIP& \OCR& & \FVD& \IS &\CLIP& \OCR& & \FVD& \IS &\CLIP& \OCR& & & \FVD& \IS &\CLIP& \OCR \\ \midrule

\textbf{Ours} & & \tbf{147.4} & \tul{1.79} & \tbf{0.87} & \tbf{0.71} & & \tbf{177.52} & \tul{1.75} & \tbf{0.88} & \tul{0.54} & & \tbf{36.16} & 1.62 & \tul{0.94} & \tbf{0.98} & & & \tbf{125.90} & 1.77 & \tbf{0.88} & \tbf{0.76} \\ \midrule

$-$Effect separation & & 789.8 & 2.75 & 0.74 & 0.49 & & 880.1 & 2.88 & 0.75 & 0.49 & & 607.1 & 2.54 & 0.82 & 0.23 & & & 729.8 & 2.73 & 0.75 & 0.48 \\

$-\mathcal{C}_{word}$  & & 239.1 & \tbf{1.77} & \tul{0.86} & \tul{0.70} & & 247.1 & \tbf{1.73} & \tul{0.87} & \tul{0.54} & & 62.7 & \tul{1.60} & \tbf{0.95} & \tul{0.95} & & & 201.8 & \tbf{1.74} & \tul{0.87} & 0.74 \\ 

$-$$L_\textit{glyph}$     & & \tul{188.9} & \tbf{1.77} & \tul{0.86} & \tul{0.70} & & \tul{226.8} & \tul{1.75} & \tbf{0.88} & \tbf{0.55} & & \tul{50.1} & \tbf{1.59} & 0.93 & \tbf{0.98} & & & \tul{160.8} & \tul{1.75} & \tbf{0.88} & \tul{0.75} \\

\bottomrule
\end{tabular}
}
\vspace{-6mm}
\label{tab:ablation}
\end{table}

\vspace{-2mm}
\subsection{Ablation Study}
\label{sec:experiment_ablation}

\noindent{\textbf{Without caption separation.}}
First of all, we train our network without dividing static and dynamic caption. The performances drop significantly for every metrics in~\cref{tab:ablation}. Since we train the text-to-image backbone using only static captions, dynamic captions are considered as noisy input. Similarly, the motion module, used to train dynamic motions, is effective when the simple motion guidance is given, compared to using static and dynamic caption together.

\noindent{\textbf{Text contents incorporation.}}
We observe that text-to-image diffusion models often fail to catch up on text contents in the prompt when a long description is given. 
We thus use $\mathcal{C}_{word}$ as an additional input to give a strong attention to each letter. As shown in~\cref{tab:ablation}, without $\mathcal{C}_{word}$, the performance degradation is observed as expected.

\noindent{\textbf{Without glyph loss.}}
Lastly, we evaluate the performance without $L_\textit{glyph}$. The inferior FVD scores come from the blur effect at the edge of the word along with color bleeding effects.

\vspace{-3mm}
\section{Conclusion}
In this paper, we propose a generative kinetic typography model, named KineTy. To achieve this, we first build a large-scale kinetic typography dataset by collecting 584 templates from professional designers, and use them to train our diffusion-based network. To effectively handle the text prompt, we split it into static and dynamic captions. They are used to directly guide the spatial and temporal cross-attention for better appearance and motion effects, respectively. Our glyph loss also strengthens the legible and artistic letter generation. We lastly demonstrate that our KineTy produces the visually compelling and semantically coherent kinetic typography videos. The extensive user studies further validate the effectiveness of our model with respect to its practical utility.

\noindent\textbf{Acknowledgements} This research was supported by `Project for Science and Technology Opens the Future of the Region' program through the INNOPOLIS FOUNDATION funded by Ministry of Science and ICT (Project Number: 2022-DD-UP-0312), the National Research Foundation of Korea(NRF) grant funded by the Korea government(MSIT)(RS-2024-00338439), GIST-MIT Research Collaboration grant funded by the GIST in 2024, and the Institute of Information $\&$ communications Technology Planning $\&$ Evaluation (IITP) grant funded by the Korea government (MSIT) (No.2019-0-01842, Artificial Intelligence Graduate School Program (GIST)).

\clearpage

\bibliographystyle{splncs04}
\bibliography{main}

\begin{thebibliography}{10}
\providecommand{\url}[1]{\texttt{#1}}
\providecommand{\urlprefix}{URL }
\providecommand{\doi}[1]{https://doi.org/#1}

\bibitem{anderson2022neural}
Anderson, D., Shamir, A., Fried, O.: Neural font rendering. arXiv preprint arXiv:2211.14802  (2022)

\bibitem{azadi2018multi}
Azadi, S., Fisher, M., Kim, V.G., Wang, Z., Shechtman, E., Darrell, T.: Multi-content gan for few-shot font style transfer. In: Proceedings of the IEEE/CVF Conference on Computer Vision and Pattern Recognition (CVPR) (2018)

\bibitem{bain2021frozen}
Bain, M., Nagrani, A., Varol, G., Zisserman, A.: Frozen in time: A joint video and image encoder for end-to-end retrieval. In: Proceedings of the IEEE/CVF International Conference on Computer Vision (ICCV) (2021)

\bibitem{barratt2018note}
Barratt, S., Sharma, R.: A note on the inception score. arXiv preprint arXiv:1801.01973  (2018)

\bibitem{berio2022strokestyles}
Berio, D., Leymarie, F.F., Asente, P., Echevarria, J.: Strokestyles: Stroke-based segmentation and stylization of fonts. ACM Transactions on Graphics (TOG)  (2022)

\bibitem{blattmann2023stable}
Blattmann, A., Dockhorn, T., Kulal, S., Mendelevitch, D., Kilian, M., Lorenz, D., Levi, Y., English, Z., Voleti, V., Letts, A., et~al.: Stable video diffusion: Scaling latent video diffusion models to large datasets. arXiv preprint arXiv:2311.15127  (2023)

\bibitem{blattmann2023align}
Blattmann, A., Rombach, R., Ling, H., Dockhorn, T., Kim, S.W., Fidler, S., Kreis, K.: Align your latents: High-resolution video synthesis with latent diffusion models. In: Proceedings of the IEEE/CVF Conference on Computer Vision and Pattern Recognition (CVPR) (2023)

\bibitem{campbell2014learning}
Campbell, N.D., Kautz, J.: Learning a manifold of fonts. ACM Transactions on Graphics (TOG)  (2014)

\bibitem{chen2023joint}
Chen, C.H., Liu, Y.T., Zhang, Z., Guo, Y.C., Zhang, S.H.: Joint implicit neural representation for high-fidelity and compact vector fonts. In: Proceedings of the IEEE/CVF International Conference on Computer Vision (ICCV) (2023)

\bibitem{chen2023textdiffuser}
Chen, J., Huang, Y., Lv, T., Cui, L., Chen, Q., Wei, F.: Textdiffuser-2: Unleashing the power of language models for text rendering. arXiv preprint arXiv:2311.16465  (2023)

\bibitem{chen2024textdiffuser}
Chen, J., Huang, Y., Lv, T., Cui, L., Chen, Q., Wei, F.: Textdiffuser: Diffusion models as text painters. Proceedings of the Neural Information Processing Systems (NeurIPS)  (2023)

\bibitem{esser2023structure}
Esser, P., Chiu, J., Atighehchian, P., Granskog, J., Germanidis, A.: Structure and content-guided video synthesis with diffusion models. In: Proceedings of the IEEE/CVF International Conference on Computer Vision (ICCV) (2023)

\bibitem{ford1997kinetic}
Ford, S., Forlizzi, J., Ishizaki, S.: Kinetic typography: issues in time-based presentation of text. In: CHI'97 extended abstracts on Human factors in computing systems, pp. 269--270. ACM Digital Library (1997)

\bibitem{fridsma2019adobe}
Fridsma, L., Gyncild, B.: Adobe After Effects CC Classroom in a Book. Adobe Press (2019)

\bibitem{fu2023neural}
Fu, B., He, J., Wang, J., Qiao, Y.: Neural transformation fields for arbitrary-styled font generation. In: Proceedings of the IEEE/CVF Conference on Computer Vision and Pattern Recognition (CVPR) (2023)

\bibitem{ge2023preserve}
Ge, S., Nah, S., Liu, G., Poon, T., Tao, A., Catanzaro, B., Jacobs, D., Huang, J.B., Liu, M.Y., Balaji, Y.: Preserve your own correlation: A noise prior for video diffusion models. In: Proceedings of the IEEE/CVF International Conference on Computer Vision (ICCV) (2023)

\bibitem{guo2023sparsectrl}
Guo, Y., Yang, C., Rao, A., Agrawala, M., Lin, D., Dai, B.: Sparsectrl: Adding sparse controls to text-to-video diffusion models. arXiv preprint arXiv:2311.16933  (2023)

\bibitem{guo2023animatediff}
Guo, Y., Yang, C., Rao, A., Wang, Y., Qiao, Y., Lin, D., Dai, B.: Animate{D}iff: Animate your personalized text-to-image diffusion models without specific tuning. arXiv preprint arXiv:2307.04725  (2023)

\bibitem{hessel2021clipscore}
Hessel, J., Holtzman, A., Forbes, M., Le~Bras, R., Choi, Y.: Clipscore: A reference-free evaluation metric for image captioning. In: Proceedings of the 2021 Conference on Empirical Methods in Natural Language Processing (2021)

\bibitem{ho2022imagen}
Ho, J., Chan, W., Saharia, C., Whang, J., Gao, R., Gritsenko, A., Kingma, D.P., Poole, B., Norouzi, M., Fleet, D.J., et~al.: Imagen video: High definition video generation with diffusion models. arXiv preprint arXiv:2210.02303  (2022)

\bibitem{hu2022make}
Hu, Y., Luo, C., Chen, Z.: Make it move: controllable image-to-video generation with text descriptions. In: Proceedings of the IEEE/CVF Conference on Computer Vision and Pattern Recognition (CVPR) (2022)

\bibitem{huang2021unifying}
Huang, Y., Xue, H., Liu, B., Lu, Y.: Unifying multimodal transformer for bi-directional image and text generation. In: Proceedings of the 29th ACM International Conference on Multimedia. pp. 1138--1147 (2021)

\bibitem{iluz2023word}
Iluz, S., Vinker, Y., Hertz, A., Berio, D., Cohen-Or, D., Shamir, A.: Word-as-image for semantic typography. ACM Transactions on Graphics (TOG)  (2023)

\bibitem{jahanian2013recommendation}
Jahanian, A., Liu, J., Lin, Q., Tretter, D., O'Brien-Strain, E., Lee, S.C., Lyons, N., Allebach, J.: Recommendation system for automatic design of magazine covers. In: Proceedings of the 2013 International Conference on Intelligent User Interfaces (IUI) (2013)

\bibitem{jia2023cole}
Jia, P., Li, C., Liu, Z., Shen, Y., Chen, X., Yuan, Y., Zheng, Y., Chen, D., Li, J., Xie, X., et~al.: Cole: A hierarchical generation framework for graphic design. arXiv preprint arXiv:2311.16974  (2023)

\bibitem{kato2015textalive}
Kato, J., Nakano, T., Goto, M.: Textalive: Integrated design environment for kinetic typography. In: Proceedings of the 33rd Annual ACM Conference on Human Factors in Computing Systems (CHI) (2015)

\bibitem{krishnan2023textstylebrush}
Krishnan, P., Kovvuri, R., Pang, G., Vassilev, B., Hassner, T.: Textstylebrush: Transfer of text aesthetics from a single example. IEEE Transactions on Pattern Analysis and Machine Intelligence (TPAMI)  (2023)

\bibitem{lee2002kinetic}
Lee, J.C., Forlizzi, J., Hudson, S.E.: The kinetic typography engine: an extensible system for animating expressive text. In: Proceedings of the 15th annual ACM symposium on User Interface Software and Technology (UIST) (2002)

\bibitem{lee2006using}
Lee, J., Jun, S., Forlizzi, J., Hudson, S.E.: Using kinetic typography to convey emotion in text-based interpersonal communication. In: Proceedings of the 6th conference on Designing Interactive systems (DIS) (2006)

\bibitem{li2021few}
Li, C., Taniguchi, Y., Lu, M., Konomi, S.: Few-shot font style transfer between different languages. In: Proceedings of the IEEE/CVF Winter Conference on Applications of Computer Vision (WACV) (2021)

\bibitem{li2023compositional}
Li, X., Wu, L., Wang, C., Meng, L., Meng, X.: Compositional zero-shot artistic font synthesis. In: Proceedings of the International Joint Conferences on Artificial Intelligence (IJCAI) (2023)

\bibitem{liu2021decoupled}
Liu, X., Meng, G., Chang, J., Hu, R., Xiang, S., Pan, C.: Decoupled representation learning for character glyph synthesis. IEEE Transactions on Multimedia (TMM)  (2021)

\bibitem{liu2022learning}
Liu, Y.T., Guo, Y.C., Li, Y.X., Wang, C., Zhang, S.H.: Learning implicit glyph shape representation. IEEE Transactions on Visualization and Computer Graphics (TVCG)  (2022)

\bibitem{liu2023dualvector}
Liu, Y.T., Zhang, Z., Guo, Y.C., Fisher, M., Wang, Z., Zhang, S.H.: Dualvector: Unsupervised vector font synthesis with dual-part representation. In: Proceedings of the IEEE/CVF Conference on Computer Vision and Pattern Recognition (CVPR) (2023)

\bibitem{loshchilov2018decoupled}
Loshchilov, I., Hutter, F.: Decoupled weight decay regularization. In: Proceedings of the International Conference on Learning Representations (ICLR) (2018)

\bibitem{luo2023videofusion}
Luo, Z., Chen, D., Zhang, Y., Huang, Y., Wang, L., Shen, Y., Zhao, D., Zhou, J., Tan, T.: Videofusion: Decomposed diffusion models for high-quality video generation. In: Proceedings of the IEEE/CVF Conference on Computer Vision and Pattern Recognition (CVPR) (2023)

\bibitem{men2019dyntypo}
Men, Y., Lian, Z., Tang, Y., Xiao, J.: Dyntypo: Example-based dynamic text effects transfer. In: Proceedings of the IEEE/CVF Conference on Computer Vision and Pattern Recognition (CVPR) (2019)

\bibitem{minakuchi2005automatic}
Minakuchi, M., Tanaka, K.: Automatic kinetic typography composer. In: Proceedings of the ACM SIGCHI International Conference on Advances in Computer Entertainment Technology (ACE) (2005)

\bibitem{mu2024font}
Mu, X., Chen, L., Chen, B., Gu, S., Bao, J., Chen, D., Li, J., Yuan, Y.: Fontstudio: Shape-adaptive diffusion model for coherent and consistent font effect generation. arXiv preprint arXiv:2406.08392  (2024)

\bibitem{nagata2023contour}
Nagata, Y., Iwana, B.K., Uchida, S.: Contour completion by transformers and its application to vector font data. arXiv preprint arXiv:2304.13988  (2023)

\bibitem{gpt4vision}
OpenAI: {GPT}-4{V}(ision) system card. \url{https://cdn.openai.com/papers/GPTV_System_Card.pdf} (2023)

\bibitem{pan2023few}
Pan, W., Zhu, A., Zhou, X., Iwana, B.K., Li, S.: Few shot font generation via transferring similarity guided global style and quantization local style. In: Proceedings of the IEEE/CVF International Conference on Computer Vision (ICCV) (2023)

\bibitem{qu2023exploring}
Qu, Y., Tan, Q., Xie, H., Xu, J., Wang, Y., Zhang, Y.: Exploring stroke-level modifications for scene text editing. In: Proceedings of the AAAI Conference on Artificial Intelligence (AAAI) (2023)

\bibitem{radford2021learning}
Radford, A., Kim, J.W., Hallacy, C., Ramesh, A., Goh, G., Agarwal, S., Sastry, G., Askell, A., Mishkin, P., Clark, J., et~al.: Learning transferable visual models from natural language supervision. In: Proceedings of the International Conference on Machine Learning (ICML) (2021)

\bibitem{reddy2021multi}
Reddy, P., Zhang, Z., Wang, Z., Fisher, M., Jin, H., Mitra, N.: A multi-implicit neural representation for fonts. Proceedings of the Neural Information Processing Systems (NeurIPS)  (2021)

\bibitem{rombach2022high}
Rombach, R., Blattmann, A., Lorenz, D., Esser, P., Ommer, B.: High-resolution image synthesis with latent diffusion models. In: Proceedings of the IEEE/CVF Conference on Computer Vision and Pattern Recognition (CVPR) (2022)

\bibitem{shimoda2021rendering}
Shimoda, W., Haraguchi, D., Uchida, S., Yamaguchi, K.: De-rendering stylized texts. In: Proceedings of the IEEE/CVF International Conference on Computer Vision (ICCV) (2021)

\bibitem{shimoda2024towards}
Shimoda, W., Haraguchi, D., Uchida, S., Yamaguchi, K.: Towards diverse and consistent typography generation. In: Proceedings of the IEEE/CVF Winter Conference on Applications of Computer Vision (WACV) (2024)

\bibitem{singer2022make}
Singer, U., Polyak, A., Hayes, T., Yin, X., An, J., Zhang, S., Hu, Q., Yang, H., Ashual, O., Gafni, O., et~al.: Make-a-video: Text-to-video generation without text-video data. In: Proceedings of the International Conference on Learning Representations (ICLR) (2023)

\bibitem{smith2012adobe}
Smith, J., Team, A.C.: Adobe After Effects CS6 Digital Classroom. John Wiley \& Sons (2012)

\bibitem{subramanian2021strive}
Subramanian, J., Chordia, V., Bart, E., Fang, S., Guan, K., Bala, R., et~al.: Strive: Scene text replacement in videos. In: Proceedings of the IEEE/CVF International Conference on Computer Vision (ICCV) (2021)

\bibitem{tanveer2023ds}
Tanveer, M., Wang, Y., Mahdavi-Amiri, A., Zhang, H.: Ds-fusion: Artistic typography via discriminated and stylized diffusion. arXiv preprint arXiv:2303.09604  (2023)

\bibitem{tham2024vecfu}
Thamizharasan, V., Liu, D., Agarwal, S., Fisher, M., Gharbi, M., Wang, O., Jacobson, A., Kalogerakis, E.: Vecfusion: Vector font generation with diffusion. In: Proceedings of the IEEE/CVF Conference on Computer Vision and Pattern Recognition (CVPR) (2024)

\bibitem{tuo2023anytext}
Tuo, Y., Xiang, W., He, J.Y., Geng, Y., Xie, X.: Anytext: Multilingual visual text generation and editing. In: Proceedings of the International Conference on Learning Representations (ICLR) (2024)

\bibitem{unterthiner2018towards}
Unterthiner, T., Van~Steenkiste, S., Kurach, K., Marinier, R., Michalski, M., Gelly, S.: Towards accurate generative models of video: A new metric \& challenges. arXiv preprint arXiv:1812.01717  (2018)

\bibitem{wang2023anything}
Wang, C., Wu, L., Liu, X., Li, X., Meng, L., Meng, X.: Anything to glyph: Artistic font synthesis via text-to-image diffusion model. In: SIGGRAPH Asia 2023 Conference Papers (2023)

\bibitem{wang2023cf}
Wang, C., Zhou, M., Ge, T., Jiang, Y., Bao, H., Xu, W.: Cf-font: Content fusion for few-shot font generation. In: Proceedings of the IEEE/CVF Conference on Computer Vision and Pattern Recognition (CVPR) (2023)

\bibitem{wang2023lavie}
Wang, Y., Chen, X., Ma, X., Zhou, S., Huang, Z., Wang, Y., Yang, C., He, Y., Yu, J., Yang, P., et~al.: Lavie: High-quality video generation with cascaded latent diffusion models. arXiv preprint arXiv:2309.15103  (2023)

\bibitem{wang2022self}
Wang, Y., Ye, Y., Mao, Y., Yu, Y., Song, Y.: Self-supervised scene text segmentation with object-centric layered representations augmented by text regions. In: Proceedings of the 30th ACM International Conference on Multimedia (2022)

\bibitem{wang2021deepvecfont}
Wang, Y., Lian, Z.: Deepvecfont: Synthesizing high-quality vector fonts via dual-modality learning. ACM Transactions on Graphics (TOG)  (2021)

\bibitem{wang2022aesthetic}
Wang, Y., Pu, G., Luo, W., Wang, Y., Xiong, P., Kang, H., Lian, Z.: Aesthetic text logo synthesis via content-aware layout inferring. In: Proceedings of the IEEE/CVF Conference on Computer Vision and Pattern Recognition (CVPR) (2022)

\bibitem{wang2023deepvecfont}
Wang, Y., Wang, Y., Yu, L., Zhu, Y., Lian, Z.: Deepvecfont-v2: Exploiting transformers to synthesize vector fonts with higher quality. In: Proceedings of the IEEE/CVF Conference on Computer Vision and Pattern Recognition (CVPR) (2023)

\bibitem{wong1996temporal}
Wong, Y.Y.: Temporal typography: a proposal to enrich written expression. In: Proceedings of the Conference Companion on Human Factors in Computing Systems (CHI) (1996)

\bibitem{xia2023vecfontsdf}
Xia, Z., Xiong, B., Lian, Z.: Vecfontsdf: Learning to reconstruct and synthesize high-quality vector fonts via signed distance functions. In: Proceedings of the IEEE/CVF Conference on Computer Vision and Pattern Recognition (CVPR) (2023)

\bibitem{xie2023creating}
Xie, L., Shu, X., Su, J.C., Wang, Y., Chen, S., Qu, H.: Creating emordle: Animating word cloud for emotion expression. IEEE Transactions on Visualization and Computer Graphics (TVCG)  (2023)

\bibitem{xie2023wakey}
Xie, L., Zhou, Z., Yu, K., Wang, Y., Qu, H., Chen, S.: Wakey-wakey: Animate text by mimicking characters in a gif. In: Proceedings of the 36th Annual ACM Symposium on User Interface Software and Technology (2023)

\bibitem{xu2023unsupervised}
Xu, C., Zhou, M., Ge, T., Jiang, Y., Xu, W.: Unsupervised domain adaption with pixel-level discriminator for image-aware layout generation. In: Proceedings of the IEEE/CVF Conference on Computer Vision and Pattern Recognition (CVPR) (2023)

\bibitem{xu2021rethinking}
Xu, X., Zhang, Z., Wang, Z., Price, B., Wang, Z., Shi, H.: Rethinking text segmentation: A novel dataset and a text-specific refinement approach. In: Proceedings of the IEEE/CVF Conference on Computer Vision and Pattern Recognition (CVPR) (2021)

\bibitem{yang2020swaptext}
Yang, Q., Huang, J., Lin, W.: Swaptext: Image based texts transfer in scenes. In: Proceedings of the IEEE/CVF Conference on Computer Vision and Pattern Recognition (CVPR) (2020)

\bibitem{yang2017awesome}
Yang, S., Liu, J., Lian, Z., Guo, Z.: Awesome typography: Statistics-based text effects transfer. In: Proceedings of the IEEE/CVF Conference on Computer Vision and Pattern Recognition (CVPR) (2017)

\bibitem{yang2020te141k}
Yang, S., Wang, W., Liu, J.: Te141k: artistic text benchmark for text effect transfer. IEEE Transactions on Pattern Analysis and Machine Intelligence (TPAMI)  (2020)

\bibitem{yang2021shape}
Yang, S., Wang, Z., Liu, J.: Shape-matching gan++: Scale controllable dynamic artistic text style transfer. IEEE Transactions on Pattern Analysis and Machine Intelligence (TPAMI)  (2021)

\bibitem{yang2019controllable}
Yang, S., Wang, Z., Wang, Z., Xu, N., Liu, J., Guo, Z.: Controllable artistic text style transfer via shape-matching gan. In: Proceedings of the IEEE/CVF International Conference on Computer Vision (ICCV) (2019)

\bibitem{yang2023glyphcontrol}
Yang, Y., Gui, D., Yuan, Y., Liang, W., Ding, H., Hu, H., Chen, K.: Glyphcontrol: Glyph conditional control for visual text generation. In: Proceedings of the Neural Information Processing Systems (NeurIPS) (2023)

\bibitem{yang2023fontdiffuser}
Yang, Z., Peng, D., Kong, Y., Zhang, Y., Yao, C., Jin, L.: Fontdiffuser: One-shot font generation via denoising diffusion with multi-scale content aggregation and style contrastive learning. arXiv preprint arXiv:2312.12142  (2023)

\bibitem{zhang2023adding}
Zhang, L., Rao, A., Agrawala, M.: Adding conditional control to text-to-image diffusion models. In: Proceedings of the IEEE/CVF International Conference on Computer Vision (ICCV) (2023)

\bibitem{zhang2023editing}
Zhang, S., Ma, J., Wu, J., Ritchie, D., Agrawala, M.: Editing motion graphics video via motion vectorization and transformation. ACM Transactions on Graphics (TOG)  (2023)

\end{thebibliography}
\end{document}